\documentclass[nohyperref]{article}

\usepackage{microtype}
\usepackage{graphicx}
\usepackage{subfigure}
\usepackage{booktabs} 

\usepackage{hyperref}

 \usepackage[accepted]{icml2022}

\usepackage{amsmath}
\usepackage{amssymb}
\usepackage{mathtools}
\usepackage{amsthm}

\usepackage[capitalize,noabbrev]{cleveref}

\theoremstyle{plain}

\theoremstyle{definition}

\theoremstyle{remark}

\usepackage{mathtools} 
\usepackage{booktabs} 
\usepackage{tikz} 
\usepackage[utf8]{inputenc} 
\usepackage[T1]{fontenc}    
\usepackage{hyperref}       
\usepackage{url}            
\usepackage{booktabs}       
\usepackage{amsfonts}       
\usepackage{nicefrac}       
\usepackage{microtype}      
\usepackage{microtype}
\usepackage{graphicx}
\usepackage{subfigure}
\usepackage{booktabs} 
\usepackage{bbm}
\usepackage{amsmath}
\usepackage{mathtools}
\usepackage{xcolor}

\usepackage{booktabs}
\usepackage{caption}
\usepackage{float}
\usepackage{titlesec}

\usepackage{capt-of}
\usepackage{float}
\usepackage{tabularx} 
\usepackage{tcolorbox}
\usepackage{grffile}
\usepackage{wrapfig, blindtext}
\usepackage{array}
\usepackage{arydshln}
\usepackage{multirow}
\usepackage{tikz}
\usetikzlibrary{automata,positioning}
\usepackage{amsfonts}
\usepackage{bm}
\usepackage{xfrac}
\usepackage{stmaryrd}
\usepackage{mathtools}
\usepackage{eucal}
\usepackage{multicol}
\usepackage{bbm}

\DeclarePairedDelimiterX{\infdivx}[2]{(}{)}{%
	#1\;\delimsize\|\;#2%
}

\DeclareMathOperator*{\argmax}{arg\,max}  
\usepackage{placeins}

\usepackage{hyperref}

\usepackage{algorithm}
\usepackage{algorithmic}

\usepackage[utf8]{inputenc} 
\usepackage[T1]{fontenc}    
\usepackage{hyperref}       
\usepackage{url}            
\usepackage{booktabs}       
\usepackage{amsfonts}       
\usepackage{nicefrac}       
\usepackage{microtype}      
\usepackage{xcolor}         
\DeclareMathAlphabet{\mathcal}{OMS}{cmsy}{m}{n}

\usepackage[textsize=tiny]{todonotes}

\icmltitlerunning{Extra Exploration with Distributional Critics}

\begin{document}

\twocolumn[
\icmltitle{Exploration with Multi-Sample Target Values \\ for Distributional Reinforcement Learning}



\icmlsetsymbol{equal}{*}

\begin{icmlauthorlist}
\icmlauthor{Michael Teng}{yyy}
\icmlauthor{Michiel van de Panne}{comp}
\icmlauthor{Frank Wood}{comp,sch}
\end{icmlauthorlist}

\icmlaffiliation{comp}{Department of Computer Science, University of British Columbia, Vancouver, Canada}
\icmlaffiliation{sch}{Montr\'eal Institute for Learning Algorithms, Montr\'eal, Canada}
\icmlaffiliation{yyy}{Department of Engineering Science, University of Oxford, Oxford, United Kingdon}
\icmlcorrespondingauthor{Michael Teng}{mteng@robots.ox.ac.uk}

\icmlkeywords{Machine Learning, ICML}

\vskip 0.3in
]



\printAffiliationsAndNotice{}  

\begin{abstract}
Distributional reinforcement learning (RL) aims to learn a value-network that predicts the full distribution of the returns for a given state, often modeled via a quantile-based critic. This approach has been successfully integrated into common RL methods for continuous control, giving rise to algorithms such as Distributional Soft Actor-Critic (DSAC). In this paper, we introduce multi-sample target values (MTV) for distributional RL, as a principled replacement for single-sample target value estimation, as commonly employed in current practice. The improved distributional estimates further lend themselves to UCB-based exploration. These two ideas are combined to yield our distributional RL algorithm, E2DC (Extra Exploration with Distributional Critics). We evaluate our approach on a range of continuous control tasks and demonstrate state-of-the-art model-free performance on difficult tasks such as Humanoid control. We provide further insight into the method via visualization and analysis of the learned distributions and their evolution during training.
\end{abstract}

\section{Introduction}


The recent success of deep reinforcement learning (RL) in continuous control tasks is largely attributable to advances in model-free actor-critic algorithms \citep{haarnoja2019soft,fujimoto2018addressing,gu2016q,lillicrap2015continuous,schulman2015high,mnih2016asynchronous,schulman2017proximal,kurutach2018model}. These approaches generally follow two main design principles: (1) a neural network parametrized estimate of the expectation of the value function (critic) and (2) a stochastic policy to encourage exploration (actor). While this set of methods have been shown to be robust, estimating the expectation of the value function is at odds with the intrinsic stochasticity of the cumulative reward ahead introduced by the policy (Figure~\ref{fig:hist}; bottom-left). To address this, the field has recently turned to modeling the full distribution of returns, introducing the `distributional RL' framework. 


Along with the inefficiency of fitting a point estimate to the empirical distribution of returns, only capturing the first moment introduces additional drawbacks for tackling exploration in the RL setting. When RL is viewed as a bandit problem, the stochastic nature of the observed return of each action at a given state has multiple sources, including the sampling policy used during training to take subsequent actions, and the stochastic nature of the observed rewards and state transitions. When the parametrized value estimate does not quantify uncertainty over the return, exploration techniques used in bandit literature cannot readily be ported to actor-critic algorithms. Recently, work has begun to address this issue by leveraging uncertainty estimates of the estimated value-function to direct exploration in the RL setting. \citet{ciosek2019better} computes an upper and lower bound using an ensemble of value-functions, \citet{chen2017ucb} utilize an ensemble of actor-critic agents to compute an upper confidence bound (UCB) for exploration, \citet{lee2020sunrise} introduce a state-of-the-art ensemble method called SUNRISE, \citet{tang2018exploration} apply posterior sampling for exploration with distributional critics, and finally, both \citet{moerland2017efficient} and \citet{mavrin2018exploration} use the uncertainty of distributional value-functions for additional exploration using Thompson sampling and UCB, respectively. 


Despite these recent advances that leverage either a distributional critic or an ensemble of critics for exploration purposes, there remains a gap in designing a simple, principled approach for additional exploration based on value-function estimation. This is particularly true for continuous control tasks and being able to achieve stronger performance on standard benchmarks. To the best of our knowledge, the exploration problem in state-of-the-art actor-critic methods for continuous control has largely been left to adding randomness to the policy, either with entropy regularization in Soft Actor-Critic (SAC; \citet{haarnoja2018soft}), adding noise to a deterministic policy in deep deterministic policy gradients (DDPG; \citet{lillicrap2015continuous}, TD3; \citet{fujimoto2018addressing}), or to use more expressive policies with hierarchies or normalizing flows \citep{haarnoja2018latent, ward2019improving}. Otherwise, recent advances to leverage critics for exploration purposes must train ensembles of both policies and their respective value-functions, which is computationally expensive and performs comparably to RL methods that are not designed for sample efficiency. This is reflected in Table~\ref{sunrise}, where the ensemble-based SUNRISE method~\citep{lee2020sunrise} performs roughly on par
with a distributional method such as IQN~\citep{dabney2018implicit}.

In this paper, we address this gap by taking advantage of the advances in distributional RL, which rely on estimating the full distribution of returns at each state. The distributional RL framework has been initially applied to discrete control tasks, and a series of improvements have been made since. First, \citet{bellemare2017distributional} introduce the C51 algorithm using a KL-divergence loss to fit a distributional estimate of the value-function. \citet{dabney2018distributional} improves upon this work by replacing the KL-divergence loss with quantile regression to learn a better estimate called Quantile Regression Deep Q-Networks (QR-DQN). Finally, \citet{dabney2018implicit} introduce an architectural improvement to the value-function estimator called Implicit Quantile Networks (IQN). Subsequently, continuous control counterparts have been used successfully, with distributed distributional DDPG (D4PG) adding QR-DQN to the DDPG algorithm \citep{barth2018distributed} and distributed SAC (DSAC) integrating IQN into the SAC algorithm \citep{ma2020distributional}. 
We continue in the direction of these works via a principled multi-sample improvement to the use of the distributional estimates, in the context of continuous control. 
We then show that these can be effectively combined with bandit-like UCB-based approaches to further exploration.

The rest of this paper is organized as follows. Section 2 reviews relevant background including the SAC, IQN, and DSAC algorithms. Section 3 introduces our novel algorithm called Extra Exploration with Distributional Critics (E2DC), which modifies the way in which IQN captures uncertainty and couples this with UCB for exploration. Section 4 gives experimental results for our algorithm and Section 5 discuses future work.

%



\section{Background}

\label{stochastic_control}
In this section, we review relevant background starting with the RL problem setup and the algorithms we build on including SAC, IQN, and DSAC. 

Formally, the RL problem considers a task modeled as a finite-time Markov decision process (MDP) comprised of states $s \in S$, actions $a \in A$, initial state distribution $p_0(s_0)$, transition function $p(s_{t+1}|s_t,a_t)$, and the random reward $R : S\times A \rightarrow \mathbb{R}$, distributed as $R(s,a) \sim r(\cdot | s, a)$. We wish to find the conditional distribution $\pi(a_t|s_t)$, referred to as the policy, that maximizes the expectation over the cumulative reward under the system dynamics $p(s_{t+1}|s_t,a_t)$ and initial conditions $p(s_0)$ under some discount factor, $\gamma$:
\begin{equation}
	 J(\pi) = \mathop{\mathbb{E}}_{\pi, p, r}\bigg[\sum_{t} \gamma^t R(a_t,s_t)\bigg]
\end{equation}
We define the value-function, $V(s)$, and the action value-function, $Q^\pi(s,a)$ (also referred to as the Q-function or Q-value). $Q^\pi(s,a)$ is defined as the discounted expectation of the sum of rewards ahead following some policy after taking some action, $a$, in some state, $s$ (this action is not necessarily following the policy). The value-function is then the Q-function marginalized over all actions for a given state. 
\begin{align}
	Q^\pi(s,a) &:= \mathop{\mathbb{E}}_{\pi, p, r}\bigg[\sum_{t} \gamma^t R(a_t,s_t)\quad |\quad a_0 = a,~s_0 = s \bigg] \\
	V^\pi(s) &:= \mathop{\mathbb{E}}_{a\sim\pi(\cdot|s)}\bigg[ Q^\pi(s,a) \bigg] 
\end{align}
Next, we define the random variable, $Z^\pi(s,a)$, whose possible values are the full support of the distribution over discounted sum of rewards ahead starting at some state-action pair, $(s,a)$, and following the policy, transition function, and random environment interactions:
\begin{align}
	Z^\pi(s,a)  &\stackrel{D}{:=} \sum_{t} \gamma^t R(a_t,s_t) \\
	|\quad a_0 &= a,~s_0 = s, s_{t + 1} \sim p(\cdot|s_t, a_t), a_t \sim \pi(\cdot | s_t),\nonumber\\
	Z^\pi(s) &\stackrel{D}{:=} Z^\pi(s,a)\quad |\quad a \sim \pi(\cdot|s)
\end{align}
where for two random variables, $U,V$, $U \stackrel{D}{:=} V$ denotes that $U$ and $V$ follow the same distribution. We write $Z(s,a)$ as distributed according to $P_Z(\cdot | s, a)$ for clarity. The goal of distributional RL is to somehow capture or estimate $P_Z^\pi$ in contrast to standard expected RL, which aims to estimate $Q^\pi$. Finally, it should be clear that the relationship between $Q$ and $Z$ is that the $Q$-value is the mean of $Z$:
\begin{equation}
	Q^\pi(s,a) := \mathbb{E}_{P_Z}[Z^\pi(s,a)] \label{eq:ez}
\end{equation}

  \begin{figure*}[tb]
  	\begin{center}
  		\includegraphics[width=0.8\textwidth]{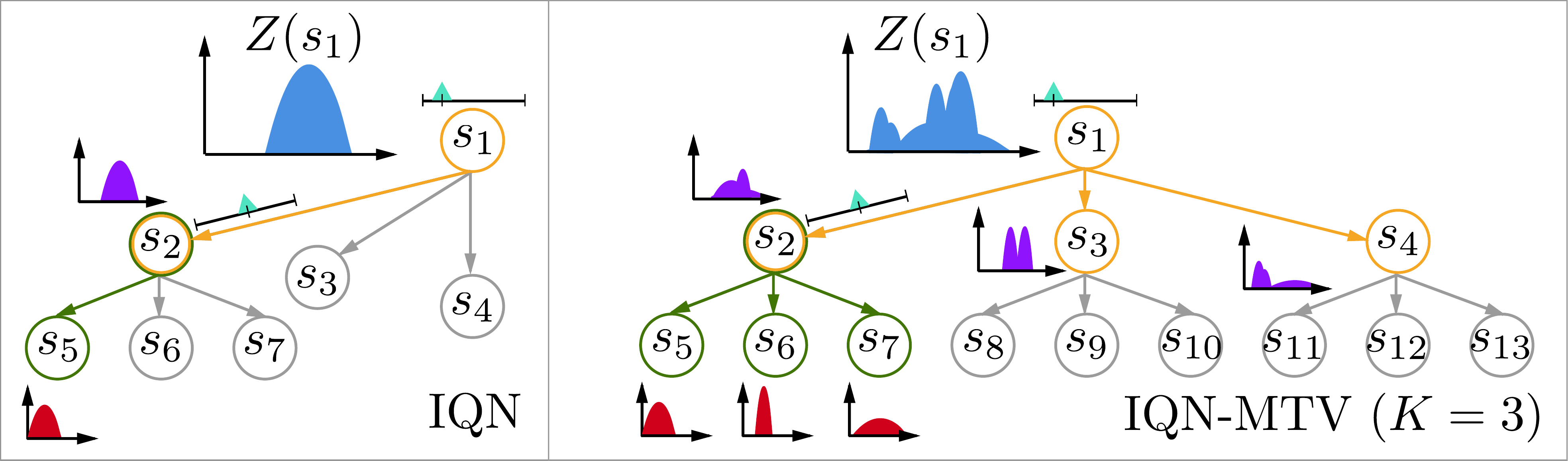}
  		\caption{Propagation of information in temporal difference Z-learning for IQN and IQN-MTV. In this illustrative MDP, there are 3 possible actions at each state: left, center, right. For each layer in the tree, the green and yellow path denotes subsequent actions that are pooled in the Bellman target. The teal triangles denote the reward obtained by arriving at states $R(s_1),R(s_2)$, which are added to a discounted Z-values from either single states in IQN or multiple states for IQN-MTV. Intuitively, higher $K$ in IQN-MTV leads to easier learning of multimodal Z-networks. } \label{fig:diagram}
  	\end{center}
  	\vskip -0.0in
 \end{figure*}

%

\subsection{Soft Actor-Critic}
Soft Actor-Critic is a state-of-the-art, model-free algorithm for continuous control. \citet{haarnoja2018soft} introduce SAC as soft policy iteration with a dueling critic architecture combined with a reparametrized stochastic policy based on the maximum entropy framework. In the following notation, we assume the entropy bonus as part of the stochastic reward at each timestep for clarity. They show that their algorithm converges to an optimal policy for maximizing the expected return subject to a maximum entropy policy by iteratively performing soft policy evaluation and soft policy improvement. 

We first define the soft policy evaluation step. For a fixed policy, $\pi$, and a initial parametrization of a Q-function, $Q_\theta$, a soft Bellman operator, $\mathcal{T}$ is defined:
	\begin{equation}
	(\mathcal{T}^\pi Q)(s,a) = R(s,a) + \gamma\mathop{\mathbb{E}}_{s'\sim p}\bigg[ V(s')\bigg]. \label{eq:classic-bellman}
\end{equation}

Because this operator is contractive in the space of possible Q-functions, i.e. $||\mathcal{T}^\pi Q_1 - \mathcal{T}^\pi Q_2||_\infty \leq \gamma|| Q_1 - Q_2||_\infty $, repeated applications of the operator to the initial $Q_\theta$ converges to the fixed point solution of $Q^\pi$ \citep{haarnoja2017reinforcement}. This leads to a gradient update which minimizes the MSE distance between the current estimate produced by $Q_\theta$ and the application of the Bellman operator, referred to as the \textit{target}. This is referred to as Q-learning, using samples from a replay buffer, $\mathcal{D}$, in the off-policy setting:
{\small\begin{align}
	&\mathcal{L}^Q_{\theta}(s,a) =\label{eq:classic-td}\\
	& \mathop{\mathbb{E}}_{s,a\sim\mathcal{D}} \bigg[ \mathop{\mathbb{E}}_{s'\sim p}\bigg[\bigg(R(s,a) + \gamma  \mathbb{E}_{a'\sim \pi}[Q_{\bar \theta}(s',a')]- Q_\theta(s,a)\bigg)^2\bigg]\bigg], \nonumber
\end{align}}
where $\bar\theta$ denotes frozen parameters (either synchronized with $\theta$ at some fixed iterations or updated using a moving average of $\theta$. \citet{haarnoja2018soft} show that learning a separate parametrized estimate of $V(s)$ can be done to stabilize training, however, in practice, the value-estimate of $s'$ in the above equation is often approximated with a single sample of the Q-function, i.e. $V(s') \approx Q(s', \mathrm{sample}\{\pi(\cdot|s')\})$.

Next, for soft policy improvement, we update the parametrized policy, $\pi_\phi$, by minimizing the expected KL-divergence between current policy and the exponential of the Q-value. The loss derived from this can use the reparametrization trick when assuming Gaussian policies:
\begin{align}
	&\mathcal{L}_\phi^{\pi}(s) = \mathop{\mathbb{E}}_{s\sim \mathcal{D}}\left[\mathop{\mathbb{E}}_{a\sim\pi_\phi(\cdot|s)}\Big[Q_\theta( a, s)+\alpha\log\pi_{\phi}(a|s) \Big]\right],
\end{align}
where $\alpha$ is a hyperparameter controlling how entropic the policy should be. Finally, more recent implementations of SAC also tunes a temperature coefficient using,
\begin{equation}
     \mathcal{L}^\alpha(s) = \mathbb{E}_{s\sim \mathcal{D}}\left[\mathbb{E}_{a\sim\pi}\left[\alpha\log\pi(a|s) - \alpha\mathcal{\bar H}\right]\right],
\end{equation}
which targets a fixed entropy, $\mathcal{\bar H}$ defined as the expected negative log of the policy $\bar\pi$, under sampled actions throughout optimization. We refer the reader to \citet{haarnoja2019soft} for details. 
\begin{table*}
  \caption{Performance of trained agents at 200K timesteps. The results shown are mean and standard deviation averaged over 5 seeds and the first two rows are taken from \citet{lee2020sunrise}.}
  \label{sunrise}
  \vskip 0.1in
  \centering
  \begin{tabular}{lllll}
    \toprule
    Easy tasks     & Cheetah     & Walker & Hopper   & Ant\\
    \midrule
    SAC & $4035.7 \pm 268.0$  & $-382.5 \pm 849.5$ & $2020.6 \pm 692.9$ & $836.5 \pm 68.4$   \\
    SUNRISE     & $5370.6 \pm 483.1$  & $ 1926.5\pm 694.8$ & $\bm{2601.9 \pm 306.5}$ & $\bm{1627.0 \pm 292.7}$       \\
    \midrule
        SAC (ours)    & $5547.2 \pm 261.2$  & $ 556.8\pm 115.8$ & $1833.7 \pm 614.8$ & $1551.5 \pm 254.0$  \\
    IQN      & $6489.7 \pm 655.8$  & $ \bm{1964.5\pm 594.8}$ & $1604.4\pm 757.7$ & $1557.6 \pm 602.9$  \\
    E2DC       & $\bm{7004.8 \pm 856.5}$  & $ 1789.9\pm 979.3$ & $1689.5 \pm 615.8$ & $1087.5 \pm 374.1$   \\
    \toprule
     Hard tasks     & Swimmer     & Humanoid & Humanoid Stand   & Biped Walker (H)\\
 \midrule
        SAC (ours)    & $41.4 \pm 0.59$  & $ 862.3\pm 149.1$ & $149$K $\pm 4.6$K & $-97.8 \pm 14.1$  \\
    IQN      & $\bm{48.9 \pm 0.81}$  & $ 1154\pm 249.4$ & $140$K $\pm 30.8$K & $-0.05 \pm 43.5$  \\
    E2DC       & $48.8 \pm 2.3$  & $ \bm{2218.3\pm 1394.5}$ & \bm{$176$}\textbf{K }\bm{$\pm 21.5$}\textbf{K} & \bm{$7.93 \pm 52.3$}   \\
    \bottomrule
  \end{tabular}
  \vskip -0.1in
\end{table*}
\subsection{Implicit Quantile Networks}

Implicit Quantile Networks is a distributional RL method developed first for discrete action spaces. In the distributional critic setting, we first define a distributional Bellman operator:
\begin{align}
	(\mathcal{T}^\pi Z)(s,a) &\stackrel{D}{:=} R(s,a) + \gamma Z(s',\argmax_{a'\in \mathcal{A}}Q(s', a')) \quad \nonumber \\
	&| \quad s' \sim p(\cdot|s,a), \label{eq:dbo}
\end{align} 
where $Q(s,a)$ is obtained via Equation~\ref{eq:ez}. In \citet{dabney2018distributional}, the authors replace mean-valued Q-functions with quantile functions, $Z^\tau$, which approximate the inverse CDF, $F^{-1}_{Z}(\tau) := \mathrm{inf}\{z\in R : \tau \leq F_Z(z)\}$, where $F_Z(z) := Pr(Z < z)$ denotes the CDF. For a given quantile fraction sampled uniformly on the interval, $\tau \sim [0,1]$, then $Z^\tau$ is a sample of the random variable $Z$. 

Given a parametrized quantile function, $Z_\theta^\tau(s,a)$, the authors show that using the Huber quantile regression loss gives a contractive distributional Bellman operator (Equation~\ref{eq:dbo}) in the $p$-Wasserstein metric for inverse CDFs, i.e.  $d_\mathcal{Z}(\mathcal{T}^\pi Z_1,\mathcal{T}^\pi Z_2) \leq \gamma d_\mathcal{Z}(Z_1,Z_2)$, where:
\begin{equation}
	d_\mathcal{Z} := W_p(U,V) = \bigg(\int_0^1 |F_U^{-1}(w) - F_V^{-1}(w)|^p dw\bigg)^{1/p},
\end{equation}
Finally, this leads to a loss that minimizes the Wasserstein distance between the parametrized quantile function, $Z_\theta$ and its Bellman target, referred to as Z-learning:
\begin{align}
	\mathcal{L}^Z_{\theta}(s,a) &= \mathop{\mathbb{E}}_{s,a\sim\mathcal{D}} \bigg[ \mathop{\mathbb{E}}_{s'\sim p}\bigg[\frac{1}{M}\sum_{i=1}^N \sum_{j=1}^M\rho^{H}_\tau(\zeta(s,a,s',a'))\bigg]\bigg],\nonumber \\ 
	\zeta(\cdot) &= R(s,a) + \gamma  Z_\theta^{\tau_j}(s', \max_{a'}Q(s',a'))- Z^{\tau_i}_\theta(s,a)\nonumber \\
	\rho^H_\tau(u) &= |\tau - \mathbbm{1}_{u < 0}|\mathcal{L}_H^\kappa(u),\nonumber\\
	\mathcal{L}_H^\kappa(u) &= 
\begin{cases}
    \frac{1}{2}u^2,& \text{if } |u| \le \kappa\\
    \kappa(|u| - \frac{1}{2}\kappa),              & \text{otherwise}
\end{cases}   \nonumber\\
\{\tau_i\}_{i=1}^{N}&, \{\tau_j\}_{j=1}^{M} \sim \mathrm{Uniform}([0,1])\label{eq:dist-td} 
\end{align}
\subsection{Distributional SAC}
In Distributional SAC, \citet{ma2020distributional} update the SAC algorithm with a quantile critic. Instead of Equation~\ref{eq:dbo}, the authors define the distributional soft Bellman operator:
\begin{align}
	(\mathcal{T}^\pi_{\mathrm{soft}} Z)(s,a) &\stackrel{D}{:=} R(s,a) + \gamma Z(s') \quad \label{eq:dsbo}\\
	&\stackrel{D}{:=} R(s,a) + \gamma \Big[Z(s',a') - \alpha \log \pi(a'|s')\Big] \quad \nonumber\\
	&\quad| \quad s' \sim p(\cdot|s,a),~a'\sim \pi(\cdot|s') \nonumber
\end{align}
In practice, Equation~\ref{eq:dsbo} is approximated with a single sample, $a' \sim \pi(\cdot|s')$, as is the case in the classic SAC algorithm. Our contribution here takes advantage of distributional critics, in that samples from multiple action-value distributions can be pooled to compute a better estimate of $Z(s)$. Additionally, DSAC retains the advances of SAC with entropy tuning and a dueling Z-network estimate, $Z^\tau := \min(Z^\tau_{\theta_2},Z^\tau_{\theta_1})$. Finally, DSAC performs policy improvement step that with an expectation of the Q-value by taking the mean of the Z-distribution


  \begin{figure*}[tb]
  	\begin{center}
  		\includegraphics[width=0.95\textwidth]{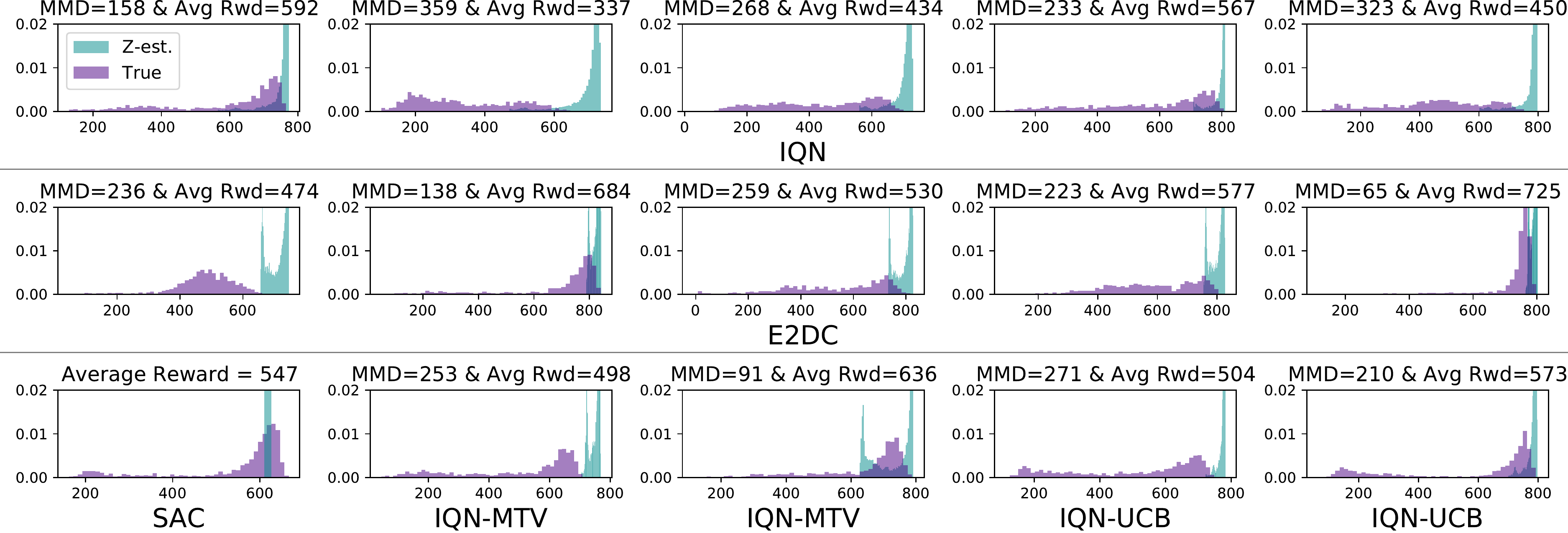}
  		\caption{Histograms of true empirical return distributions following a trained policy (purple) compared against distributions given by a trained Z-network (cyan) on the Humanoid-v2 tasks. Each plot is a independent actor-critic pair. We highlight results from the baseline algorithms, SAC and IQN, along with our algorithm, E2DC, and its ablations, IQN-UCB and IQN-MTV. In the first two rows, we show 5 histogram plots each for IQN and E2DC, respectively, and in the third row, we show one example for SAC, followed by two each for IQN-MTV and IQN-UCB.   } \label{fig:hist}
  	\end{center}
  	\vskip -0.1in
 \end{figure*}


\section{Extra Exploration with Distributional Critics}
 
In this section, we introduce our method for additional exploration, but first introduce two concepts that motivated our algorithm. First, we distinguish between uncertainty quantified by using ensembles of critics and distributional critics. The former relies on capturing parametric uncertainty, which is uncertainty over the parameters given in the beginning of training, while the latter relies on capturing return uncertainty, i.e. the uncertainty of returns based on a stochastic policy. \citet{moerland2017efficient} compare using distributional critics for exploration using return uncertainty against using Bayesian neural networks for exploration leveraging parametric uncertainty. In 4 of the 5 difficult tasks they considered, exploration based on the return uncertainty outperformed all others. In this work, we do not focus on ensembles or Bayesian neural networks to capture parametric uncertainty. 

Secondly, we note that the uncertainty associated with the return distribution is also subject to bootstrapping error in temporal difference settings. \citet{kumar2019stabilizing} showed that when using a target value-function in the start of training, the estimates given by the target estimate will be off due to sampling actions outside of the behavior policy. Without being able to explore these actions, the errors themselves are propogated throughout the Q-function, since they are TD steps. Furthermore, prior work has shown that this error commonly manifests itself as over-estimation \citep{van2016deep, wang2016dueling, lan2020maxmin}. Taken to the distributional setting, the uncertainty estimates given by distributional critics would likely have higher variance when encountering out of distribution actions. As a result, distributional methods also capture \textit{parametric} uncertainty through the accumulation of these errors, especially so at the start of training \citep{mavrin2018exploration}.

Next, we introduce two novel methodological improvements to the DSAC algorithm. When taken together, these changes comprise our novel algorithm which we refer to as Extra Exploration with Distributional Critics (E2DC).

\subsection{Multi-sample Target Value}

In the original formulation in the SAC algorithm, the soft Bellman operator is defined for a value function, $V(s)$, which can either be learned as a separate parametrized value network distinct from the action-value network, $Q(s,a)$, or it can be approximated with a single sample from the policy and passed through the Q-function. Similarly, the value-distribution, $Z(s)$, is almost always approximated with a single sample, i.e. $Z(s) \approx Z(s,\mathrm{sample}\{\pi(\cdot|s)\})$. 

In the distributional case, we present a multi-sample approximation of $Z(s)$, obtained by ancestrally resampling from the policy and then the conditional action-value distribution. 
In this setup, our value-distribution can instead be approximated by the empirical distribution:
\begin{align}
	a_{1:K}\sim\pi(\cdot|s),~ Z_{1:M}^k \sim Z_\theta(s,a_k)  \nonumber\\
	Z(s) :=	\frac{1}{MK}\sum_{j=1}^M\sum_{k=1}^K\delta_{Z_j^{k}}
\end{align}
Using the above equation, we introduce the final loss used in the multi-sample quantile regression:
\begin{align}
	\mathcal{L}^{\mathrm{MTV}}_{\theta}(s,a) &= \mathop{\mathbb{E}}_{s,a\sim\mathcal{D}} \bigg[ \mathop{\mathbb{E}}_{s'\sim p}\bigg[\frac{1}{MK}\sum_{i=1}^N \sum_{j=1}^M\sum_{k=1}^K\rho^{H}_\tau(\zeta)\bigg]\bigg],\nonumber\\
	\zeta &= R(s,a) + \gamma  Z_\theta^{\tau_j^k}(s', a_k')- Z^{\tau_i}_\theta(s,a)
\end{align}
where $a_k'\sim \pi(\cdot|s')$, $\{\tau_j^k\}_{j=1}^{M} \sim \mathrm{Uniform}([0,1])\quad\forall k \in 1...K$, and $\{\tau_i\}_{i=1}^{N} \sim \mathrm{Uniform}([0,1])$. 
Note that in the case where $K=1$, we recover the single sample estimate of the original IQN algorithm. We call this a Multi-sample Target Value (MTV) method.  

Figure~\ref{fig:diagram} provides a schematic comparison of IQN (equivalently, single-sample MTV) and multi-sample MTV. 
Intuitively, the additional samples contribute to parametric uncertainty in the propagation of Bellman update errors at the start of training, 
and leading to a better approximation of the intrinsic return uncertainty at the end of training. 


\subsection{UCB Exploration}
Our method here extends the work of \citet{tang2018exploration}, which uses posterior sampling with distributional critics. We improve upon this by leveraging a distributional critic for UCB, similar to the way in which \citet{chen2017ucb} replaces posterior sampling used in \citet{osband2016deep} with UCB exploration for further improvements. We extend the idea of continuous action space UCB exploration from \citet{lee2020sunrise}. We call this extension IQN-UCB.

While exploring, we maintain an UCB for each action at a given state based on the current Z-distribution, $\{Z^{\tau_i}\}_{i=1}^{N} \sim Z_\theta(s,a)$. We define the empirical mean $\tilde\mu(s,a) := \frac{1}{N}\sum_i Z^{\tau_i}$ and the empirical standard deviation $\tilde\sigma(s,a) := \sqrt{ \frac{1}{N}\sum_i (Z^{\tau_i} - \tilde\mu)^2}$. Following \citet{lee2020sunrise}, we use this in the continuous control setting by first sampling $L$ candidate actions for a given state, $a^{1:L}_t \sim \pi(\cdot | s_t)$. We then choose the action from the candidate set that maximizes the UCB, $a_t = \argmax_{a^l_t} \{\tilde\mu(s_t,a_t^l) + \lambda \tilde\sigma(s_t, a_t^l)\}$, where $\lambda$ is a hyperparameter to control the importance of each action's distributional critic variance. 

  \begin{algorithm}[H]
    \caption{Extra Exploration with Distributional Critics}
	\label{alg:krac}
	\begin{algorithmic}[1]	
		\STATE {\bfseries Input:} $\theta_1$, $\theta_2$, $\phi$, $\bar{\theta}_1 \gets \theta_1$, $\bar{\theta}_2 \gets \theta_2$, $D \gets \{\} $, Hyperparameters $\alpha, \lambda, \bar\tau, N,M,K,L$, \\Learning rates $\lambda_\alpha, \lambda_\phi, \lambda_\theta$
		\WHILE{$n \leq \mathrm{max~timesteps}$}
		\FOR{ each environment step }
		\STATE sample candidate actions $a^{1:L}_t \sim \pi_\phi(a_t|s_t)$
				\STATE select action to maximize the UCB $a_t = \argmax_{a^l_t} \{\tilde\mu(s_t,a_t^l) + \lambda \tilde\sigma(s_t, a_t^l)\}$
		\STATE do update $s_{t+1} \gets$ $p(s_{t+1}|s_{t}, a_t)$, receive reward, $r_t \gets r(s_{t}, a_{t})$
		\STATE store transition in replay buffer $D \gets D \cup \{(a_t, s_t, r_t, s_{t+1})\}$
		\ENDFOR
		\FOR{ each gradient step }
		\STATE sample minibatch from replay buffer $D$
		\STATE  $\phi \gets \phi - \lambda_\phi\nabla_\phi \mathcal{L}^\pi_\phi$ using $Q(s,a) = \mathbb{E}[Z_\theta(s,a)]$
		\STATE  $\theta_{i} \gets \theta - \lambda_\theta\nabla_\theta \mathcal{L}^{\mathrm{MTV}}_{\theta_i}$ using $N,M,K$ samples drawn from $Z_\theta$, Bellman targets, and $\pi_\phi$
		\STATE $\bar \theta_i \gets (1 - \bar\tau)\bar \theta_i + \bar\tau \theta_i $, update EMA of target networks for $i\in\{1,2\}$
		\STATE  $\alpha \gets \alpha - \nabla_\alpha\lambda_\alpha\mathcal{L}^\alpha $, adjust entropy coefficient towards task-specific target entropy 
		\ENDFOR
  		\ENDWHILE
	\end{algorithmic}
  \end{algorithm}
\subsection{Algorithm Description}
\begin{figure}[tb]
\centering
\begin{minipage}[b]{0.45\linewidth}
\centering
\includegraphics[width=\linewidth]{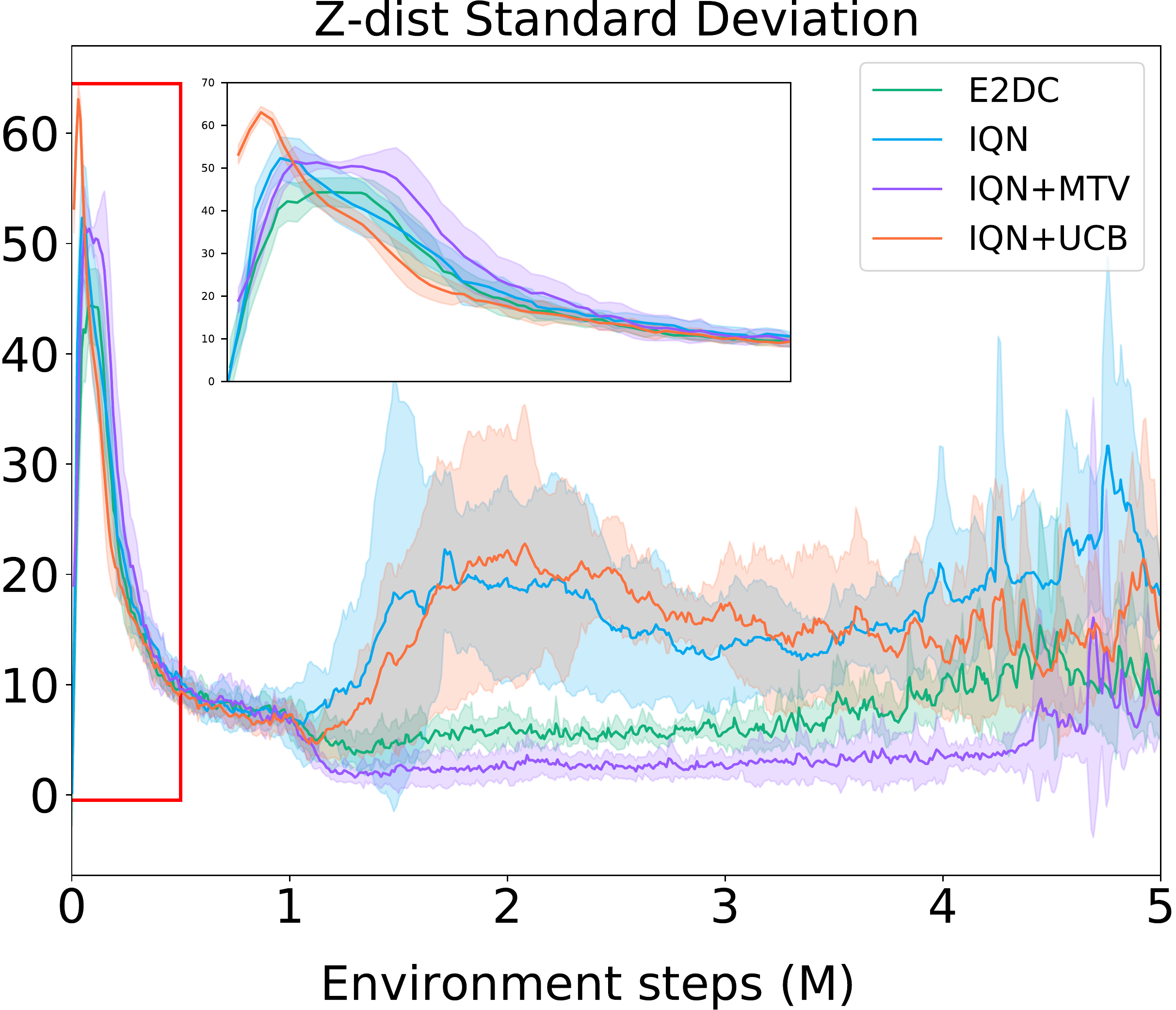}
\label{fig:minipage1}
\end{minipage}
\quad
\begin{minipage}[b]{0.45\linewidth}
\centering
\includegraphics[width=\linewidth]{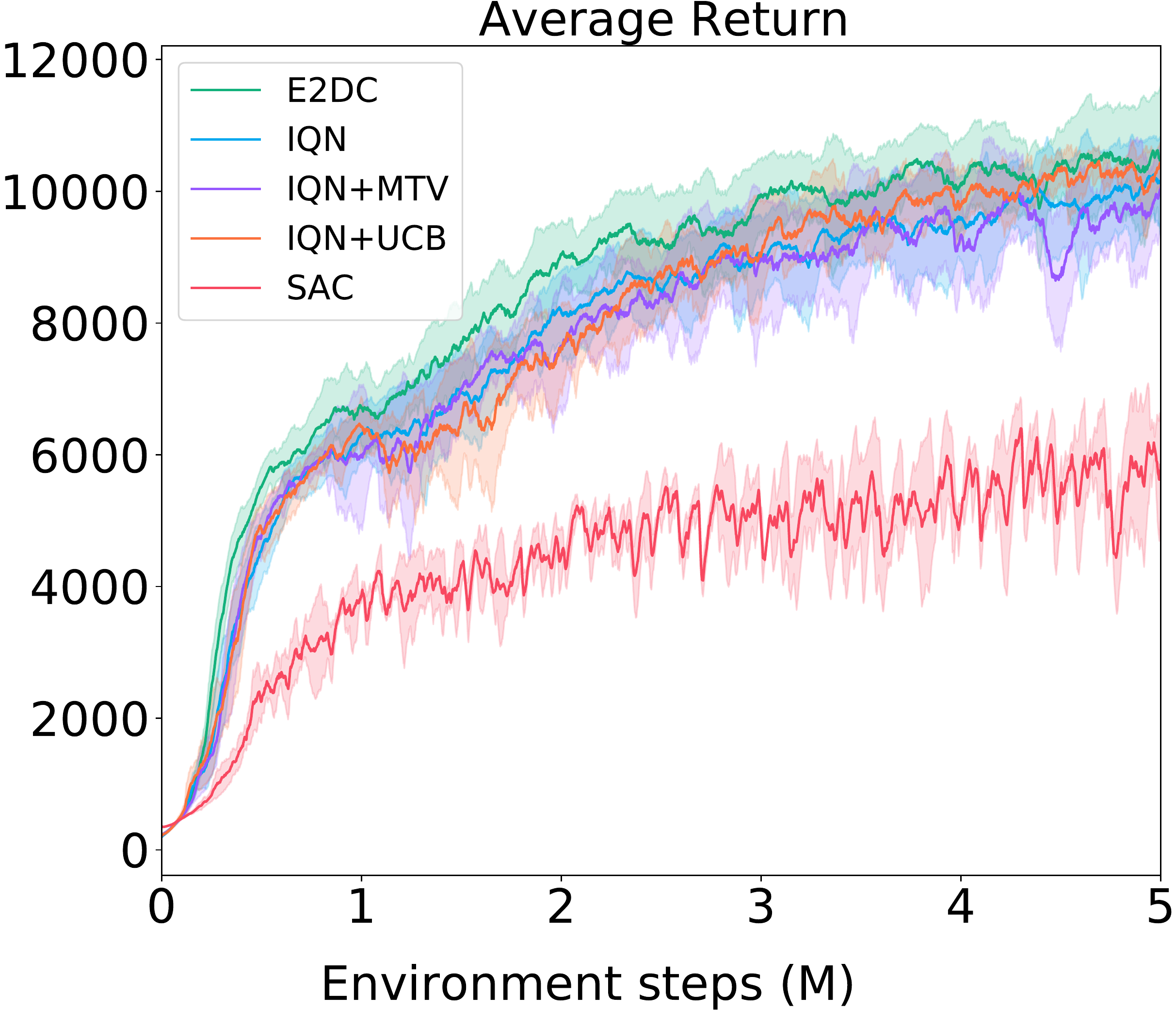}
\label{fig:minipage2}
\end{minipage}
\caption{Left: Z-distribution standard deviations throughout training for IQN, E2DC, and ablations, IQN-MTV and IQN-UCB smoothed using a 0.5 EMA. The red box is zoomed in on the first 500K timesteps. Right: average returns of each method along with SAC baseline using the deterministic policy rollout for evaluation at each timestep throughout training. } \label{fig:human}
\vskip -0.1in
\end{figure}

Finally, we outline our algorithm E2DC, which adds the two improvements to the DSAC algorithm. The first is the MTV sample estimate which adds diversity to the Bellman targets. During UCB exploration, we exactly want to target the action-values that have been fitted to these higher variance Z-distributions to explore further, as they are the ones with the highest parametric uncertainty in the start, but decreasing as training progresses. Furthermore we empirically validate this in the Humanoid task. While UCB itself does help, adding the MTV improves the stability of learning (Figure~\ref{fig:human}). 

Our algorithm is similar to DSAC in implementation, which uses two parameterized Z-networks, $\theta_1$ and $\theta_2$ and their respective targets, $\bar\theta_{1,2}$. We use the minimum of the Z-samples for the gradient estimates and update the targets using a $\bar\tau$ parameter to control the exponential moving average (EMA) update of $\theta_{1,2}$ throughout optimization. The full pseudocode is given in Algorithm~\ref{alg:krac}.

  \begin{figure*}[tb]
  	\begin{center}
  		\includegraphics[width=0.9\textwidth]{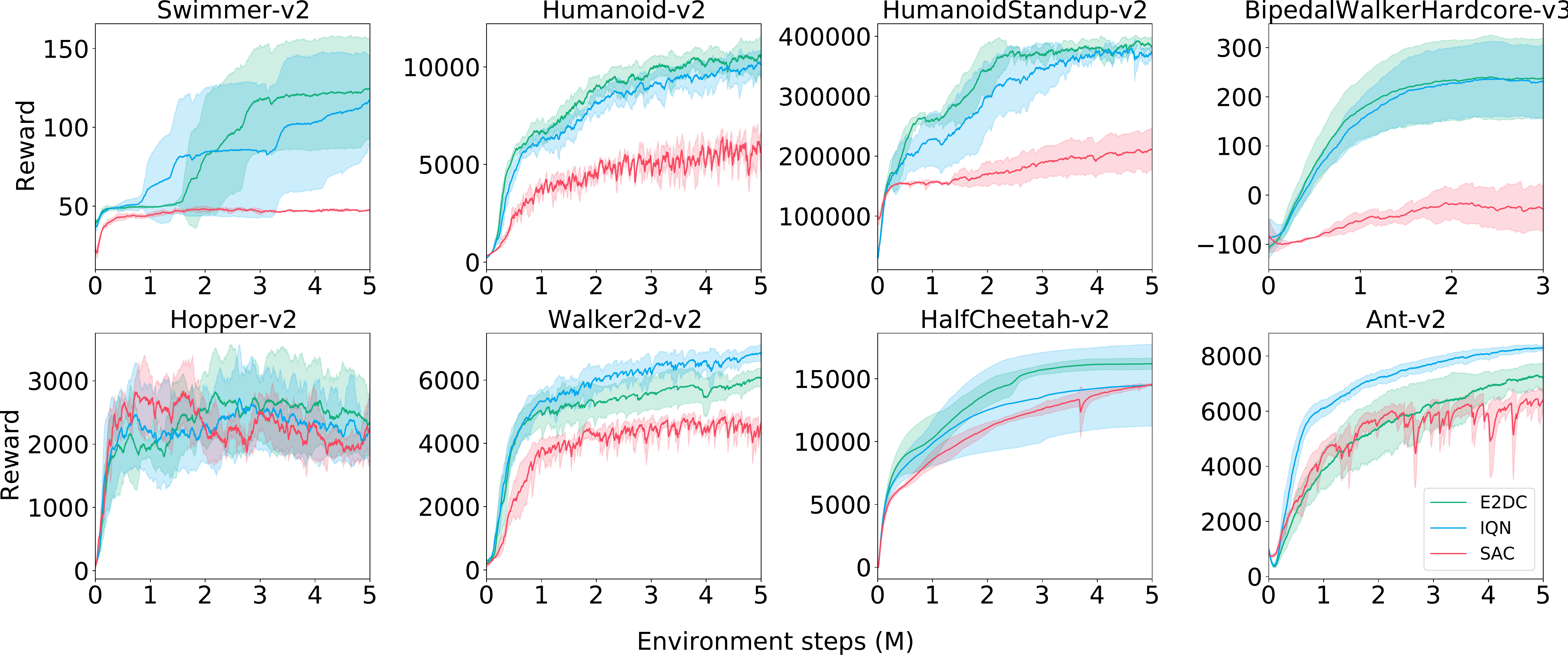}
  		\includegraphics[width=0.9\textwidth]{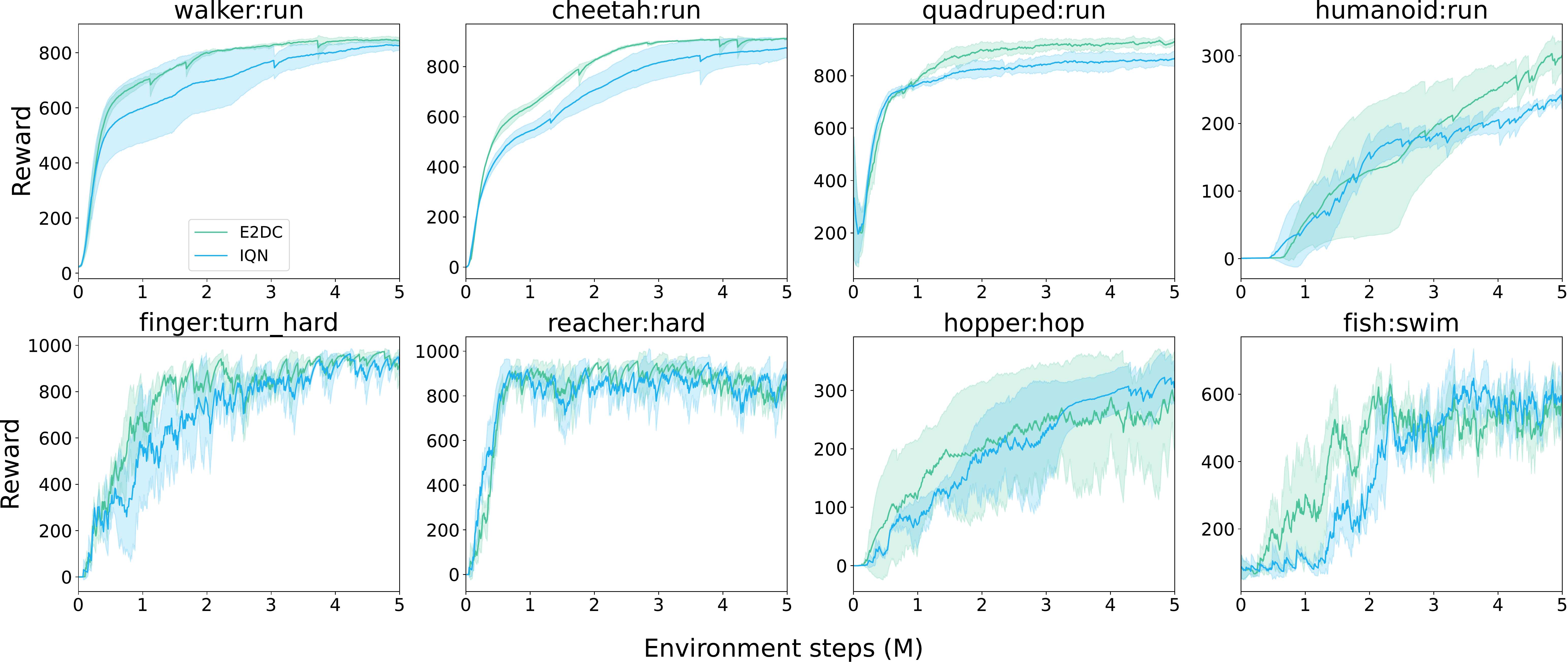}
  		\caption{Performance on MuJoCo control suite tasks (v2) and the most difficult Box3d/DeepMind control tasks. The top row are the 4 most difficult tasks for SAC, while the bottom row comprise easier tasks. For the first two rows, Humanoid-v2 IQN and E2DC use 10 seeds (others use 5). For the remaining two rows, we use 3 seeds per algorithm. } \label{fig:summary}
  	\end{center}
  	\vskip -0.1in
 \end{figure*}

\section{Experiments}
We test the effectiveness of E2DC on a host of benchmark tasks in continuous control and compare against both conventional and distributional state-of-the-art RL algorithms. The baseline distributional RL algorithm we use is DSAC with an IQN loss for Z-learning, and we refer to it as IQN in our experiments for clarity. The baseline classical RL algorithm we compare against is SAC. An implementation of both baselines is provided in the code released by \citet{ma2020distributional}, based on RLKit \citep{pongrlkit}, which we further modify to implement E2DC. For all training runs using IQN-MTV or E2DC, we set $M=8,K=8$, while runs using IQN or IQN-UCB, we set $M=64,K=1$ for a fair comparison. All runs use $N=64$, and UCB settings $L=12$ and $\lambda=50$. We note that higher values for $K$ tend to increase performance, but incur a slight computational cost.


We perform experiments using the MuJoCo suite \citep{todorov2012mujoco}, DeepMind control suite \citep{tassa2020dmcontrol}, and Box2d tasks from OpenAI Gym \citep{brockman2016openai}. All experiments are run with 3 seeds for SAC and 5 seeds for distributional methods, IQN and E2DC, unless otherwise stated. Reward curves are plotted with one standard deviation and smoothed using a 0.98 weighted EMA, unless otherwise stated. All MuJoCo/DeepMind agents are trained to 5 million timesteps and all Box2d agents are trained to 3 million timesteps. Additional implementation and experimental details are provided in the supplementary material.


\subsection{Distribution Matching Study}
In Table~\ref{tab:emd}, we quantify the 1-Wasserstein Distance (i.e. Earth Mover Distance; EMD), between true returns generated by a trained policy at 5 million timesteps and those generated by the corresponding trained Z-network on actor-critic pairs learning on Humanoid-v2. For a comparison, we fix the initial state and action sampled by the policy and rollout 500 times independently under the stochastic policy and environment transition. Therefore, the returns garnered with 500 rollouts should approximate that which is captured by the Z-network for the same initial state-action pair. We find that IQN-MTV and E2DC provide the closest matched distributions, while leading to higher average returns under the stochastic policy. Meanwhile, IQN is the worst performing algorithm when evaluating the average return given by a fully trained RL agent. 

Figure~\ref{fig:hist} provides a visualization of these policies, with additional tasks summarized in the supplementary material. We qualitatively note that the true returns under the policy rollouts across all algorithms, including SAC, tend to be more bimodal than unimodal, while only the multisample variants, E2DC and IQN-MTV, are able to capture the bimodal behavior in estimating the returns.  

We find that evaluation of these agents on the Humanoid-v2 task under the stochastic policy 
closes the gap in performance from SAC to the distributional RL algorithms. 
This is in contrast to the Humanoid-v2 results given in Figures~\ref{fig:summary} and \ref{fig:human}, 
which use a deterministic policy to evaluate the total reward, without discounting or entropy bonus.
In Figure~\ref{fig:hist} we see for a SAC agent (bottom-left) that while sometimes the policy achieves very low return, 
most of the rollouts are within the higher return mode to the right. 
In contrast, IQN agents (top row) also achieve low return at times, but tends to have less mass concentrated its the higher return region, despite much higher deterministic returns.


\begin{table*}
  \caption{Row 1: Wasserstein distance between the true distribution of returns following a trained stochastic policy, $\pi$, and the distribution given by a corresponding trained Z-network to approximate $Z^\pi$. Row 2: Average returns for different trained agents following the stochastic policy from the start.}
  \label{tab:emd}
  \vskip 0.1in
  \centering
  \begin{tabular}{llllll}
    \toprule
     ~& SAC & IQN     & IQN-UCB & IQN-MTV   & E2DC\\
    \midrule
    $\mathrm{EMD}~(W_p~;~p=1)$&- & $265 \pm 75$  & $228 \pm 36$ & \bm{$164 \pm 75$} & $183 \pm 72$   \\
  Average Return & $519 \pm 42$  & $480 \pm 98$ & $551 \pm 40$ & $575 \pm 63$ & \bm{$600 \pm 93$} \\  
    \bottomrule
  \end{tabular}
\end{table*}

\subsection{Continuous Control Study}

In Table~\ref{sunrise} and Figure~\ref{fig:summary}, we present results on the full MuJoCo Suite as well as the hardest Box2d task and DeepMind control tasks. 
The top row is comprised of the more difficult tasks, as measured by the dimensionality of the state and action spaces in the Humanoid-v2 and HumanoidStandup-v2 tasks as well as by the difficulty for SAC to solve these tasks.
We note that these achieve a performance gain in final convergence and sample efficiency when using E2DC over all baselines. However, in the second row among the easier tasks, we find that while IQN and E2DC both outperform the SAC baseline, E2DC does not improve upon IQN in Walker2d-v2 and Ant-v2.
Finally, in the bottom two rows comprises the DeepMind control suite tasks, we find that our algorithm consistently performs as well or better than the IQN baseline. 

We provide a brief intuition for why E2DC is beneficial on harder tasks. We hypothesize that this phenomenon can be attributed to easier tasks being more readily solved using only the entropy bonus for exploration. We know quantitatively, the final entropy coefficient, $\alpha$, is about 2-5x times higher for each of the easier tasks than for Humanoid-v2. Additionally, both Humanoid tasks and the Swimmer-v2 task are thought to require more exploration \citep{gangwani2018learning}. When maximum-entropy RL is enough to solve the task reasonably well, using E2DC may hinder performance by focusing computational resources on exploring when there is less need. Finally, this is reflected in the sometimes improved performance of IQN over a multi-sample bound, as the former is also more `exploitative' in Z-learning.

\subsection{Humanoid Ablation Study}
In Figure~\ref{fig:human}, we evaluate the two ablations, IQN-MTV and IQN-UCB, of E2DC on the Humanoid-v2 task. We find that for this task, E2DC still outperforms IQN and either ablation in average returns (right). In the left plot, we show the variance of each ablation and the baselines' Z-estimate throughout training. We find that the variance of the IQN-MTV Z-estimate remains highest in the initial stages of training, which we attribute to initial parametric uncertainty due to accumulation of Bellman errors. 

We therefore posit that the additional methods we introduce accomplish the goal we want in terms of exploration using return uncertainty. In the MTV method, we see that the Z-distribution remains higher variance in the beginning of training over IQN for a short period, before converging to much lower variance towards the end of training. IQN, however, has equally high variance throughout training as the IQN algorithm with UCB exploration, except towards the end (similarly reflected in equal performance between IQN and IQN-UCB on average return, except at the end). This suggests that the E2DC method helps the most, as it explores best throughout training. We believe this is because UCB targets the highly uncertain actions produced by the MTV estimator only in the beginning of training, while converging better to the true return uncertainty later in training. When IQN-MTV is then used with UCB in E2DC, these initial uncertain states are better explored, leading to higher sample efficiency and overall return shown in the right plot.

\section{Discussion}
\label{discussion}
In this work, we have introduced the multi-sample target values using pooling of quantile-based networks and leveraged this mechanism for additional exploration with UCB. Our resulting algorithm, E2DC, shows performance gains in difficult continuous control tasks and improvements in distribution matching. Future work includes integrating our changes with orthogonal advances in distributional RL \citep{kuznetsov2020controlling} and more expressive policies \citep{yue2020implicit,ward2019improving}. Due to our method potentially learning more accurate distributions of the true returns, our work here can be leveraged for specific use cases, such as risk-seeking policies in stock market trading strategies, or risk-averse learning for robotics. As with other advances in RL, the societal impact of this work is largely dependent on the RL practitioner and should be used with care in  safety-critical applications. 	

%
\bibliography{paper}

\begin{thebibliography}{35}
\providecommand{\natexlab}[1]{#1}
\providecommand{\url}[1]{\texttt{#1}}
\expandafter\ifx\csname urlstyle\endcsname\relax
  \providecommand{\doi}[1]{doi: #1}\else
  \providecommand{\doi}{doi: \begingroup \urlstyle{rm}\Url}\fi

\bibitem[Barth-Maron et~al.(2018)Barth-Maron, Hoffman, Budden, Dabney, Horgan,
  Tb, Muldal, Heess, and Lillicrap]{barth2018distributed}
Gabriel Barth-Maron, Matthew~W Hoffman, David Budden, Will Dabney, Dan Horgan,
  Dhruva Tb, Alistair Muldal, Nicolas Heess, and Timothy Lillicrap.
\newblock Distributed distributional deterministic policy gradients.
\newblock \emph{arXiv preprint arXiv:1804.08617}, 2018.

\bibitem[Bellemare et~al.(2017)Bellemare, Dabney, and
  Munos]{bellemare2017distributional}
Marc~G Bellemare, Will Dabney, and R{\'e}mi Munos.
\newblock A distributional perspective on reinforcement learning.
\newblock In \emph{International Conference on Machine Learning}, pages
  449--458. PMLR, 2017.

\bibitem[Brockman et~al.(2016)Brockman, Cheung, Pettersson, Schneider,
  Schulman, Tang, and Zaremba]{brockman2016openai}
Greg Brockman, Vicki Cheung, Ludwig Pettersson, Jonas Schneider, John Schulman,
  Jie Tang, and Wojciech Zaremba.
\newblock Openai gym.
\newblock \emph{arXiv preprint arXiv:1606.01540}, 2016.

\bibitem[Chen et~al.(2017)Chen, Sidor, Abbeel, and Schulman]{chen2017ucb}
Richard~Y Chen, Szymon Sidor, Pieter Abbeel, and John Schulman.
\newblock Ucb exploration via q-ensembles.
\newblock \emph{arXiv preprint arXiv:1706.01502}, 2017.

\bibitem[Ciosek et~al.(2019)Ciosek, Vuong, Loftin, and
  Hofmann]{ciosek2019better}
Kamil Ciosek, Quan Vuong, Robert Loftin, and Katja Hofmann.
\newblock Better exploration with optimistic actor critic.
\newblock In \emph{Advances in Neural Information Processing Systems}, pages
  1787--1798, 2019.

\bibitem[Dabney et~al.(2018{\natexlab{a}})Dabney, Ostrovski, Silver, and
  Munos]{dabney2018implicit}
Will Dabney, Georg Ostrovski, David Silver, and R{\'e}mi Munos.
\newblock Implicit quantile networks for distributional reinforcement learning.
\newblock In \emph{International conference on machine learning}, pages
  1096--1105. PMLR, 2018{\natexlab{a}}.

\bibitem[Dabney et~al.(2018{\natexlab{b}})Dabney, Rowland, Bellemare, and
  Munos]{dabney2018distributional}
Will Dabney, Mark Rowland, Marc Bellemare, and R{\'e}mi Munos.
\newblock Distributional reinforcement learning with quantile regression.
\newblock In \emph{Proceedings of the AAAI Conference on Artificial
  Intelligence}, volume~32, 2018{\natexlab{b}}.

\bibitem[Fujimoto et~al.(2018)Fujimoto, Hoof, and
  Meger]{fujimoto2018addressing}
Scott Fujimoto, Herke Hoof, and David Meger.
\newblock Addressing function approximation error in actor-critic methods.
\newblock In \emph{International Conference on Machine Learning}, pages
  1587--1596. PMLR, 2018.

\bibitem[Gangwani et~al.(2018)Gangwani, Liu, and Peng]{gangwani2018learning}
Tanmay Gangwani, Qiang Liu, and Jian Peng.
\newblock Learning self-imitating diverse policies.
\newblock \emph{arXiv preprint arXiv:1805.10309}, 2018.

\bibitem[Gu et~al.(2016)Gu, Lillicrap, Ghahramani, Turner, and Levine]{gu2016q}
Shixiang Gu, Timothy Lillicrap, Zoubin Ghahramani, Richard~E Turner, and Sergey
  Levine.
\newblock Q-prop: Sample-efficient policy gradient with an off-policy critic.
\newblock \emph{arXiv preprint arXiv:1611.02247}, 2016.

\bibitem[Haarnoja et~al.(2017)Haarnoja, Tang, Abbeel, and
  Levine]{haarnoja2017reinforcement}
Tuomas Haarnoja, Haoran Tang, Pieter Abbeel, and Sergey Levine.
\newblock Reinforcement learning with deep energy-based policies, 2017.

\bibitem[Haarnoja et~al.(2018{\natexlab{a}})Haarnoja, Hartikainen, Abbeel, and
  Levine]{haarnoja2018latent}
Tuomas Haarnoja, Kristian Hartikainen, Pieter Abbeel, and Sergey Levine.
\newblock Latent space policies for hierarchical reinforcement learning.
\newblock In \emph{International Conference on Machine Learning}, pages
  1851--1860. PMLR, 2018{\natexlab{a}}.

\bibitem[Haarnoja et~al.(2018{\natexlab{b}})Haarnoja, Zhou, Abbeel, and
  Levine]{haarnoja2018soft}
Tuomas Haarnoja, Aurick Zhou, Pieter Abbeel, and Sergey Levine.
\newblock Soft actor-critic: Off-policy maximum entropy deep reinforcement
  learning with a stochastic actor.
\newblock In \emph{International Conference on Machine Learning}, pages
  1861--1870. PMLR, 2018{\natexlab{b}}.

\bibitem[Haarnoja et~al.(2018{\natexlab{c}})Haarnoja, Zhou, Hartikainen,
  Tucker, Ha, Tan, Kumar, Zhu, Gupta, Abbeel, and Levine]{haarnoja2019soft}
Tuomas Haarnoja, Aurick Zhou, Kristian Hartikainen, George Tucker, Sehoon Ha,
  Jie Tan, Vikash Kumar, Henry Zhu, Abhishek Gupta, Pieter Abbeel, and Sergey
  Levine.
\newblock Soft actor-critic algorithms and applications, 2018{\natexlab{c}}.

\bibitem[Kumar et~al.(2019)Kumar, Fu, Tucker, and Levine]{kumar2019stabilizing}
Aviral Kumar, Justin Fu, George Tucker, and Sergey Levine.
\newblock Stabilizing off-policy q-learning via bootstrapping error reduction.
\newblock \emph{arXiv preprint arXiv:1906.00949}, 2019.

\bibitem[Kurutach et~al.(2018)Kurutach, Clavera, Duan, Tamar, and
  Abbeel]{kurutach2018model}
Thanard Kurutach, Ignasi Clavera, Yan Duan, Aviv Tamar, and Pieter Abbeel.
\newblock Model-ensemble trust-region policy optimization.
\newblock \emph{arXiv preprint arXiv:1802.10592}, 2018.

\bibitem[Kuznetsov et~al.(2020)Kuznetsov, Shvechikov, Grishin, and
  Vetrov]{kuznetsov2020controlling}
Arsenii Kuznetsov, Pavel Shvechikov, Alexander Grishin, and Dmitry Vetrov.
\newblock Controlling overestimation bias with truncated mixture of continuous
  distributional quantile critics.
\newblock In \emph{International Conference on Machine Learning}, pages
  5556--5566. PMLR, 2020.

\bibitem[Lan et~al.(2020)Lan, Pan, Fyshe, and White]{lan2020maxmin}
Qingfeng Lan, Yangchen Pan, Alona Fyshe, and Martha White.
\newblock Maxmin q-learning: Controlling the estimation bias of q-learning.
\newblock \emph{arXiv preprint arXiv:2002.06487}, 2020.

\bibitem[Lee et~al.(2020)Lee, Laskin, Srinivas, and Abbeel]{lee2020sunrise}
Kimin Lee, Michael Laskin, Aravind Srinivas, and Pieter Abbeel.
\newblock Sunrise: A simple unified framework for ensemble learning in deep
  reinforcement learning.
\newblock \emph{arXiv preprint arXiv:2007.04938}, 2020.

\bibitem[Lillicrap et~al.(2015)Lillicrap, Hunt, Pritzel, Heess, Erez, Tassa,
  Silver, and Wierstra]{lillicrap2015continuous}
Timothy~P Lillicrap, Jonathan~J Hunt, Alexander Pritzel, Nicolas Heess, Tom
  Erez, Yuval Tassa, David Silver, and Daan Wierstra.
\newblock Continuous control with deep reinforcement learning.
\newblock \emph{arXiv preprint arXiv:1509.02971}, 2015.

\bibitem[Ma et~al.(2020)Ma, Zhang, Xia, Zhou, Yang, and
  Zhao]{ma2020distributional}
Xiaoteng Ma, Qiyuan Zhang, Li~Xia, Zhengyuan Zhou, Jun Yang, and Qianchuan
  Zhao.
\newblock Distributional soft actor critic for risk sensitive learning.
\newblock \emph{arXiv preprint arXiv:2004.14547}, 2020.

\bibitem[Mavrin et~al.(2018)Mavrin, Yao, Kong, et~al.]{mavrin2018exploration}
Borislav Mavrin, Hengshuai Yao, Linglong Kong, et~al.
\newblock Exploration using distributional rl and ucb.
\newblock 2018.

\bibitem[Mnih et~al.(2016)Mnih, Badia, Mirza, Graves, Lillicrap, Harley,
  Silver, and Kavukcuoglu]{mnih2016asynchronous}
Volodymyr Mnih, Adria~Puigdomenech Badia, Mehdi Mirza, Alex Graves, Timothy
  Lillicrap, Tim Harley, David Silver, and Koray Kavukcuoglu.
\newblock Asynchronous methods for deep reinforcement learning.
\newblock In \emph{International conference on machine learning}, pages
  1928--1937, 2016.

\bibitem[Moerland et~al.(2017)Moerland, Broekens, and
  Jonker]{moerland2017efficient}
Thomas~M Moerland, Joost Broekens, and Catholijn~M Jonker.
\newblock Efficient exploration with double uncertain value networks.
\newblock \emph{arXiv preprint arXiv:1711.10789}, 2017.

\bibitem[Osband et~al.(2016)Osband, Blundell, Pritzel, and
  Van~Roy]{osband2016deep}
Ian Osband, Charles Blundell, Alexander Pritzel, and Benjamin Van~Roy.
\newblock Deep exploration via bootstrapped dqn.
\newblock \emph{arXiv preprint arXiv:1602.04621}, 2016.

\bibitem[Pong et~al.()Pong, Dalal, Lin, and Nair]{pongrlkit}
Vitchyr Pong, M~Dalal, S~Lin, and A~Nair.
\newblock Rlkit: Reinforcement learning framework and algorithms implemented in
  pytorch, 2019.
\newblock \emph{URL https://github. com/vitchyr/rlkit}.

\bibitem[Schulman et~al.(2015)Schulman, Moritz, Levine, Jordan, and
  Abbeel]{schulman2015high}
John Schulman, Philipp Moritz, Sergey Levine, Michael Jordan, and Pieter
  Abbeel.
\newblock High-dimensional continuous control using generalized advantage
  estimation.
\newblock \emph{arXiv preprint arXiv:1506.02438}, 2015.

\bibitem[Schulman et~al.(2017)Schulman, Wolski, Dhariwal, Radford, and
  Klimov]{schulman2017proximal}
John Schulman, Filip Wolski, Prafulla Dhariwal, Alec Radford, and Oleg Klimov.
\newblock Proximal policy optimization algorithms, 2017.

\bibitem[Tang and Agrawal(2018)]{tang2018exploration}
Yunhao Tang and Shipra Agrawal.
\newblock Exploration by distributional reinforcement learning.
\newblock \emph{arXiv preprint arXiv:1805.01907}, 2018.

\bibitem[Tassa et~al.(2020)Tassa, Tunyasuvunakool, Muldal, Doron, Liu, Bohez,
  Merel, Erez, Lillicrap, and Heess]{tassa2020dmcontrol}
Yuval Tassa, Saran Tunyasuvunakool, Alistair Muldal, Yotam Doron, Siqi Liu,
  Steven Bohez, Josh Merel, Tom Erez, Timothy Lillicrap, and Nicolas Heess.
\newblock Deepmind control: Software and tasks for continuous control, 2020.

\bibitem[Todorov et~al.(2012)Todorov, Erez, and Tassa]{todorov2012mujoco}
Emanuel Todorov, Tom Erez, and Yuval Tassa.
\newblock Mujoco: A physics engine for model-based control.
\newblock In \emph{2012 IEEE/RSJ International Conference on Intelligent Robots
  and Systems}, pages 5026--5033. IEEE, 2012.

\bibitem[Van~Hasselt et~al.(2016)Van~Hasselt, Guez, and Silver]{van2016deep}
Hado Van~Hasselt, Arthur Guez, and David Silver.
\newblock Deep reinforcement learning with double q-learning.
\newblock In \emph{Thirtieth AAAI conference on artificial intelligence}, 2016.

\bibitem[Wang et~al.(2016)Wang, Schaul, Hessel, Hasselt, Lanctot, and
  Freitas]{wang2016dueling}
Ziyu Wang, Tom Schaul, Matteo Hessel, Hado Hasselt, Marc Lanctot, and Nando
  Freitas.
\newblock Dueling network architectures for deep reinforcement learning.
\newblock In \emph{International conference on machine learning}, pages
  1995--2003, 2016.

\bibitem[Ward et~al.(2019)Ward, Smofsky, and Bose]{ward2019improving}
Patrick~Nadeem Ward, Ariella Smofsky, and Avishek~Joey Bose.
\newblock Improving exploration in soft-actor-critic with normalizing flows
  policies.
\newblock \emph{arXiv preprint arXiv:1906.02771}, 2019.

\bibitem[Yue et~al.(2020)Yue, Wang, and Zhou]{yue2020implicit}
Yuguang Yue, Zhendong Wang, and Mingyuan Zhou.
\newblock Implicit distributional reinforcement learning.
\newblock \emph{arXiv preprint arXiv:2007.06159}, 2020.

\end{thebibliography}

\end{document}